\documentclass[10pt,twocolumn,letterpaper]{article}

\usepackage{wacv}      

\usepackage{graphicx}
\usepackage{amsmath}
\usepackage{amssymb}
\usepackage{booktabs}
\usepackage{enumitem}
\usepackage{soul}

\usepackage{bbm}
\usepackage{pgffor}
\usepackage{tabularray}
\usepackage{float}
\usepackage{etoolbox}
\usepackage[export]{adjustbox}
\usepackage{pgfplots}
\usepgfplotslibrary{fillbetween}
\usepgfplotslibrary{groupplots}

\DeclareMathOperator*{\argmin}{arg\,min}

\pgfplotsset{compat=1.18}

\usepackage[pagebackref,breaklinks,colorlinks]{hyperref}

\usepackage[capitalize]{cleveref}
\crefname{section}{Sec.}{Secs.}
\Crefname{section}{Section}{Sections}
\Crefname{table}{Table}{Tables}
\crefname{table}{Tab.}{Tabs.}

\def\wacvPaperID{1616} 
\def\confName{WACV}
\def\confYear{2024}

\begin{document}

\title{Patch-Based Deep Unsupervised Image Segmentation using Graph Cuts}

\author{
Isaac Wasserman\\
University of Pennsylvania\\
Philadelphia, Pennsylvania\\
{\tt\small isaacrw@seas.upenn.edu}
\and
Jeov\'a Farias Sales Rocha Neto\\
Bowdoin College\\
Brunswick, Maine\\
{\tt\small j.farias@bowdoin.edu}
}
\maketitle

\begin{abstract}
  Unsupervised image segmentation aims at grouping different semantic patterns in an image without the use of human annotation. Similarly, image clustering searches for groupings of images based on their semantic content without supervision. Classically, both problems have captivated researchers as they drew from sound mathematical concepts to produce concrete applications. With the emergence of deep learning, the scientific community turned its attention to complex neural network-based solvers that achieved impressive results in those domains but rarely leveraged the advances made by classical methods. In this work, we propose a patch-based unsupervised image segmentation strategy that bridges advances in unsupervised feature extraction from deep clustering methods with the algorithmic help of classical graph-based methods. We show that a simple convolutional neural network, trained to classify image patches and iteratively regularized using graph cuts, naturally leads to a state-of-the-art fully-convolutional unsupervised pixel-level segmenter. Furthermore, we demonstrate that this is the ideal setting for leveraging the patch-level pairwise features generated by vision transformer models. Our results on real image data demonstrate the effectiveness of our proposed methodology.
\end{abstract}

\section{Introduction}
Image segmentation has long been one of the main tasks in computer vision and it has been widely applied in general image understanding or as a preprocessing step for other tasks such as object detection. It aims at corresponding each pixel in an image to a semantically relevant class, in such a way that similar pixels are assigned to the same class. This problem finds various industrial applications such as autonomous driving, medical image analysis, video surveillance and virtual reality to name a few \cite{minaee2021image}.

On the supervised front, deep learning approaches using Convolutional (CNN) and Fully Convolutional Neural Networks (FCN) achieved unprecedented results in image segmentation, as illustrated by the UNet \cite{ronneberger2015u} and DeepLab \cite{chen2017deeplab} models. Recently, however, methods using transformer-based models, such as Segformer\cite{xie2021segformer}, DETR \cite{carion2020end} and DINO \cite{caron2021emerging}, are slowly outperforming established CNN solutions. This has prompted the recent interest in deep models that utilize image patches instead of their full-sized counterparts \cite{tolstikhin2021mlp, han2022vision}, leading some to postulate that patch representations are the main source of the transformer's success  \cite{trockman2022patches}.

These accomplishments come, however, at the cost of long training schemes and the need for larger amounts of annotated data, which hinder their application in many domains where data can be expensive or scarce, such as in biology, or astrophysics \cite{yu2018methods}. These issues can be resolved via the application of unsupervised techniques instead. In this setting, one aims at creating a model that automatically discovers semantically important visual features or groups that characterize the various objects in a scene. Classically, this could be approached via variational, statistical, and graphical methods, exemplified in active contours \cite{chan2001active}, conditional random fields \cite{krahenbuhl2011efficient}, and graph cuts \cite{shi2000normalized, boykov2001interactive}. Within the deep learning literature, prominent advances were made in the field of unsupervised deep image clustering \cite{ren2022deep}, which eventually led to developments in deep image segmentation techniques \cite{kim2020unsupervised, hamilton2021unsupervised, lin2021deep, hamilton2022unsupervised, wang2022tokencut}.

In this work, we introduce GraPL, an unsupervised image segmentation technique that draws inspiration from the success of CNNs for imaging tasks, the learning strategies of deep clustering methods, and the regularization power of graph cut algorithms. Here, we alternate the training of a CNN classifier on image patches and the minimization of a clustering energy via graph cuts. To the best of our knowledge, this is the first attempt in both the deep clustering and image segmentation literature to make use of graphs cuts to solve a deep learning-based unsupervised task. We show that our zero-shot approach detects visual segments in an image without onerous unsupervised training on an entire image dataset, automatically finds a satisfactory number of image segments, and easily translates the patch-level training to efficient pixel-level inference. Furthermore, because of its structure, it also naturally incorporates pretrained patch embeddings \cite{caron2021emerging, oquab2023dinov2}, without relying on them for the final product. Finally, we show that this simple approach achieves state-of-the-art results in deep unsupervised image segmentation, demonstrating the potential of graph cuts to improve other patch-based deep segmentation algorithms. 
Specifically, we make the following main contributions with our work:
\begin{itemize}[noitemsep, topsep=0pt]
    \item We introduce GraPL, an unsupervised segmentation method that learns a fully convolutional segmenter directly from the image's patches, using an iterative algorithm regularized by graph cuts.
    \item  We show that this framework naturally employs patch embeddings for pixel-level segmentation without the need for postprocessing schemes such as CRF refining.
    \item We demonstrate that GraPL is able to outperform the state-of-the-art in deep unsupervised segmentation by iteratively training a small, low complexity CNN.
\end{itemize}

\section{Related Work}

\subsection{Deep Clustering}
With the advancements in deep supervised image classification techniques, interest in deep architectures to solve unsupervised problems followed naturally.
This pursuit led to the task of partitioning image datasets into clusters using deep representations without human supervision, inaugurating the body of work which is now referred to as  ``deep clustering.'' The interested reader is referred to \cite{ren2022deep} for a comprehensive review on the available approaches to deep clustering.

In GraPL, we treat image patches as individual images to be clustered as a pretext task to train our segmenter and efficiently use graph cuts to impose constraints on the patch clusters. To the best of our knowledge, our method is the first to use MRF-based algorithms for clustering CNN-generated visual features.


\subsection{Deep Unsupervised Image Segmentation}
As deep clustering aims to learn visual features and groupings without human annotation via deep neural models, deep unsupervised image segmentation hopes to use the same models to learn coherent and meaningful image regions without the use of ground-truth labels. To do so, many methods explore strategies that resemble those from deep clustering. Cho \textit{et al.} \cite{cho2021picie} iteratively employs $k$-means to cluster pixel-level features extracted from a network trained on different photometrically and geometrically varying image views. The work in \cite{ji2019invariant} efficiently extends a mutual information-based deep clustering algorithm to the pixel-level by recognizing that such a process can be achieved via convolution operations. 
\cite{hwang2019segsort} computes pixel embeddings from a metric learning network and segments each image using a spherical $k$-means clustering algorithm.

In \cite{kim2020unsupervised}, the authors train an FCN with pseudo-labels generated by the same network in a prior step. They attain reliable segmentations by proposing a complex loss function that ensures the similarity among pixels in shared segments, while encouraging their spacial continuity and limiting their total count. Our method, while similarly training an FCN, works on patches  and is able to reinforce spacial continuity and low segment count via our graph cut approach. Furthermore, due to its use of graph cuts, GraPL is able to naturally incorporate pairwise patch relationships. Finally, while other patch-based unsupervised solutions require a segmentation refinement stage after a patch feature clustering step   \cite{hamilton2022unsupervised, wang2022tokencut, wang2023cut}, we both discover and instill patch knowledge interactively, without the need for postprocessing our result. 


\subsection{Graph Cuts for Image Segmentation}
Modeling image generation as a Markov random field (MRF) has a long history in Computer Vision, dating its initial theoretical and algorithmic achievements to the works of Abel \textit{et al} \cite{abend1965classification} and Besag \cite{besag1986statistical}. 
Soon enough, MRFs found applications in various image processing tasks, such as edge detection, image denoising, segmentation, and stereo \cite{li2009markov}. In particular, the works conducted by Boykov and Jolly \cite{boykov2001interactive} and Boykov \textit{et al.} \cite{boykov2001fast} demonstrated that one can apply efficient min \textit{st}-cut based algorithms to solve image segmentation by modeling it as a Maximum \textit{a Posteriori} (MAP) estimator of an MRF. Their groundbreaking results made possible the emergence of classical graph-based segmentation methodologies such as GrabCut \cite{rother2004grabcut} and was, more recently, used to improve the training of CNN-based segmenters \cite{marin2019beyond}. CRF modeling, closely related to MRF, has also played an important role in refining coarse network predictions in recent segmentation methods \cite{zheng2015conditional, chen2017deeplab, chen2017rethinking}. 

In some ways, our proposed method draws inspirations from the methodologies proposed by Rother \textit{et al.} \cite{rother2004grabcut}, and Marin \textit{et al.} \cite{marin2019beyond}. In \cite{rother2004grabcut}, the authors propose GrabCut, an algorithm that iteratively bound-optimizes a segmentation energy, requiring the solution of a min \textit{st}-cut problem at each iteration in order to perform unsupervised regularized fitting of the image's appearance, which is modeled as a Gaussian mixture model. Our algorithm also uses min \textit{st}-cut solvers iteratively, but here we (1) work on patch data, instead of individual pixels, and (2) fit the image appearance using a CNN classifier. Due to the nature of CNNs, our network can seamlessly translate the patch-level classifier into a pixel-level image segmenter. In \cite{marin2019beyond}, the authors show how to perform weakly-supervised CNN segmentation via an optimizer that alternates between solving an MAP-MRF problem and gradient computation. In contrast, our method solves a fully unsupervised segmentation problem and does not use our MAP solution to adjust gradient directions.


\section{Methodology}

    \begin{figure*}
        \centering
        \includegraphics[width=\linewidth]{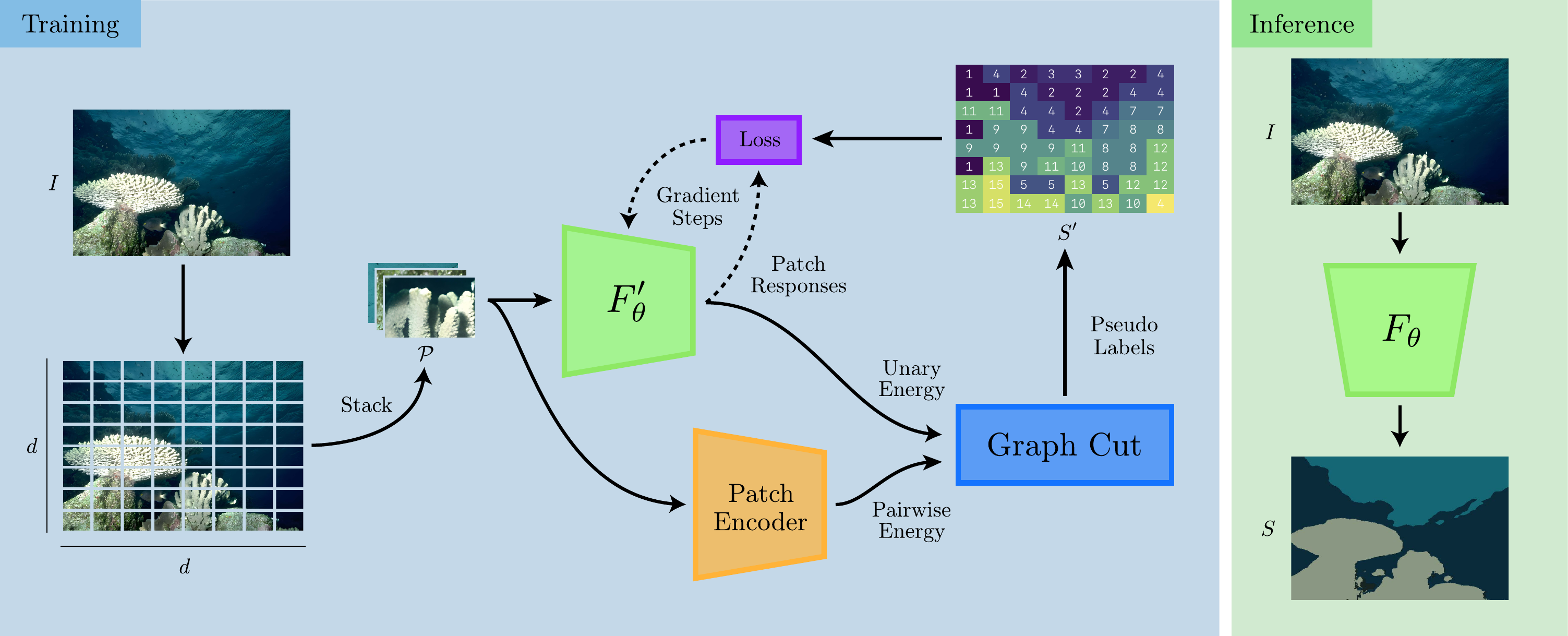}
        \caption{The proposed algorithm. GraPL trains a convolutional neural network to cluster patches of a single image without supervision under the guidance of graph cuts, spatial continuity loss, and a patch affinity encoder. At inference, this patch clustering knowledge is applied to pixel-level segmentation of the image. $F'_\theta$ and $F_\theta$ share the same parameters.}
        \label{fig:grapl_diagram}
    \end{figure*}

    GraPL (\underline{Gra}ph Cuts at the \underline{P}atch \underline{L}evel) is a fully unsupervised segmentation method which operates in a single-image paradigm. Using patch clustering as a pretext task for segmentation, during training it is able to learn distinctive segment features that enable it to effectively segment the image at the pixel level.  Although other techniques have previously shown patch clustering to be an effective surrogate task \cite{ji2019invariant, ouali2020autoregressive, wang2022tokencut, wang2023cut}, our method demonstrates that clustering the patches of a single image provides sufficient feature learning to accurately segment it. GraPL's training is guided by patch-level graph cuts; this intervention imposes spatial coherence priors which are helpful for learning clusters that are conducive to segmentation. At inference, the complexities of the pipeline disappear, leaving only the network. Leveraging a generalization of CNNs, the trained model is ``convolved'' over the entire image to produce a pixel-level segmentation.

    \setlength{\belowdisplayskip}{7pt} \setlength{\belowdisplayshortskip}{7pt}
    \setlength{\abovedisplayskip}{7pt} \setlength{\abovedisplayshortskip}{7pt}

    \subsection{Algorithm}

    Let $I: \Omega \mapsto \mathbb{R}^c$ be an image of $c$ channels with pixel set $\Omega = \{1, \ldots, n\} \times \{1, \ldots, m\}$, and $S: \Omega \mapsto \{1, \ldots, K\}$ be a segmentation map of $I$ in $K$ regions. Let $\mathcal{P}$ be a set of patches from $I$, such that all patches are of the same size, i.e., for each $p$ in $\mathcal{P}$,  $p: \Omega_p \mapsto \mathbb{R}^c, \Omega_p \subset \Omega, |\Omega_p| = \operatorname{const}$. In practice, we populate $\mathcal{P}$ by selecting all patches on a non-overlapping $d \times d$ grid of $I$, resulting in patches of shape $(w/d, h/d)$. We make this choice of $\mathcal{P}$ based on two factors: (1) efficiency, as this operation can be efficiently performed by most deep learning libraries via their unfolding methods, and (2) simplicity, as it's one of the simplest ways to generate equal sized patches that span $\Omega$. 
    
    Let $F_\theta: \Omega \mapsto [0, 1]^{K\times |\Omega|}$ be an FCN, and $F'_\theta: \Omega_p \mapsto [0, 1]^K$ be a CNN patch classifier. In GraPL, both networks are parametrized by the same parameters $\theta$. $F'_\theta$ is used in our training stage and is applied to the patches in $\mathcal{P}$, while $F_\theta$ is employed in our inference phase and is our final segmenter. The full algorithm is shown in Figure \ref{fig:grapl_diagram}. 

    \paragraph{Training Stage} Our goal is to learn $\theta$ exclusively from the data in $\mathcal{P}$ and transfer it to $F_\theta$. To do so, our method trains $F'_\theta$ by minimizing an energy formulated at the patch level of $I$. Let $S': \mathcal{P} \mapsto \{1, \ldots, K\}$ be a labeling for the patches in $\mathcal{P}$. Following the literature on MRF modeling \cite{boykov2001interactive}, we define the energy of $S'$ for an unknown $\theta$ as:
    \begin{equation}\label{equation:energy}
        E(S', \theta | \mathcal{P}) = U(S', \theta| \mathcal{P}) + \lambda V(S'|\mathcal{P}),
    \end{equation}
    with $\lambda \geq 0$. The unary term $U(\cdot)$ is traditionally defined as:
    \begin{align}\label{equation:total_unary_energy}
        U(S', \theta|X) = & -\sum_{p \in P}\sum_{k = 1}^K \mathbbm{1}(S'(p) = k) [\ln{F'_\theta(p)}]_k,
    \end{align}
    where $\mathbbm{1}(\cdot)$ is the indicator function and $[\cdot]_k$ is the $k$-th position of a vector. Let $\alpha$ and $\beta$ be probability distributions in $\mathbb{R}^K$, and let $H(\alpha, \beta) = -\sum_{k=1}^K [\alpha]_k \ln [\beta]_k$ be their cross entropy. This means that Eq. \ref{equation:total_unary_energy} can be seen as the sum of cross entropies $H(S'(p), F'_\theta(p))$ between $S'(p)$, taken as a one-hot probability distribution, and $F'_\theta(p)$ over all $p \in \mathcal{P}$. The pairwise energy term $V(\cdot)$ is given by:
    \begin{equation}\label{equation:total_pairwise_energy}
        V(S'|\mathcal{P}) = \sum_{(p, q) \in \mathcal{P}\times \mathcal{P}} \mathbbm{1}(S'(p) \neq S'(q))\phi(p, q),
    \end{equation}
    with the patch similarity function $\phi(\cdot)$ defined as: 
    \begin{equation}\label{equation:pairwise_energy}
        \phi(p,q) = \frac{1}{\text{dist}(p,q)}\exp\left(-\frac{\text{aff}(p,q)^2}{2 \sigma}\right),
    \end{equation}
    where the data affinity function $\text{aff}(p, q)$ evaluates the data similarity between $p$ and $q$, and $\text{dist}(p,q)$ considers the Euclidean distance between the centers of $p$ and $q$. We select $\sigma$ as the standard deviation of affinities for all $p, q \in \mathcal{P}$. To compute patch affinities we make use of a patch encoder, which extracts an embedding from each patch in $\mathcal{P}$. 
    
    Inspired by GrabCut \cite{rother2004grabcut}, GraPL minimizes $E$ using a block-coordinate descent iterative strategy, where we alternate between optimizing for $\theta$ and $S'$, keeping the other constant. The current labeling  $S'_{t}$ is updated using the current network parameters $\theta_{t-1}$, now taken as fixed in Eq. \ref{equation:energy}:
    \begin{equation}
        S'_{t} = \argmin_{S'} E(S' | \theta_{t-1},  \mathcal{P}).
    \end{equation}
    The above problem can be approximately solved by a series of minimum $st$-cut in the form of $\alpha$-expansion or $\alpha\beta$-swap moves \cite{boykov2001fast}. This step is can be quickly accomplished due to the efficiency of such graph cut algorithms and the comparatively small size of $\mathcal{P}$, which presents a further advantage to our patch-based framework. We then compute the updated parameters $\theta_{t}$ via:
    \begin{equation}
        \theta_{t} = \argmin_\theta \mathcal{L}(F'_\theta(\mathcal{P}), S'_{t}),
    \end{equation}
    where $F'_\theta(\mathcal{P}) = \{F'_\theta(p)\}_{p \in \mathcal{P}}$. We employ traditional gradient descent-based backpropagation to solve the above problem. The loss $\mathcal{L}(\cdot)$, is designed to predict the outputs of $F'_\theta$ on each patch using the labels from $S'_{t}$. Keeping $S'_{t}$ fixed, Eq. \ref{equation:total_unary_energy} conveniently formulates that process as a sum of cross entropy losses, just as one would naturally devise in a supervised segmentation learning scheme. We then follow Kim \etal \cite{kim2020unsupervised} and include a patch-level continuity loss $C(\theta)$: 
    \begin{equation}\label{equation:continuity_loss}
        C(\theta) = \sum_{p \in \mathcal{P}} \sum_{q \in \mathcal{N}_p} |F'_\theta(p) - F'_\theta(q)|,
    \end{equation}
    where $\lvert \cdot \rvert$ is the $L1$ norm and $\mathcal{N}_p$ is the set of patches immediately neighboring $p$ in $\Omega$ space. In the general case, one can employ a $k$-nearest neighbors graph of the elements in $\mathcal{P}$, considering the Euclidean distance between patch centers. For the $d \times d$ grid from Figure \ref{fig:grapl_diagram}, we choose $\mathcal{N}_p$ to be given by the patches immediately above and to the left of $p$, resembling what is done in \cite{kim2020unsupervised}. This continuity loss brings spatial coherence outside the graph step and encourages smooth boundaries on the network outputs. In practice, we found it to be beneficial to have both the graph step and $C(\theta)$ in our method. Our final loss is then defined as:
        \begin{equation}\label{equation:nn_loss}
          \mathcal{L}(F'_\theta(\mathcal{P}), S'_{t-1})= \sum_{p \in P}H(S_{t-1}'(p), F'_{\theta}(p)) + \mu C(\theta),
        \end{equation}
    where $\mu \geq 0$. As a consequence of the use of both graph cuts and the continuity loss described above, GraPL naturally suppresses extraneous labels arising from irrelevant patterns or textures, automatically promoting model selection.    As the alternation continues, $F'_{\theta}$ improves to the point where it no longer requires the guidance of the graph cuts to produce spatially and semantically coherent patch clusters. At that point, we end our training phase. 

    \paragraph{Inference Stage} Once $F'_\theta$ is trained, our next goal is to classify all possible patches in $I$ of shape equal to the patches in $\mathcal{P}$. To that end, we first assume that, as a CNN, $F'_\theta$ is composed of an initial series of convolutional layers and a final stage of say $Q$ dense layers, along with a softmax function at the end. Assume that the inputs of all layers are unpadded, and that each dense layer has $s_q$ units leading to a final output of size $K$. Now, one can replace each dense layer in $F'_\theta$ with a convolutional one of kernel size $\sqrt{s_q}$ and retain its exact functionality. Our resulting FCN, $F_\theta$, is now capable of efficiently being applied to $I$, by effectively ``convolving'' it with patch classifier $F'_\theta$.

    \subsection{Advantages of using Graph Cut Iterations}
    In the absence of labels, GraPL learns to cluster patches via an iterative procedure. This general formulation allows us to inject knowledge about the domain by designing an apt method for selecting pseudo-labels. While similar methods use $k$-means \cite{caron2018deep}, mixture models \cite{hwang2019segsort}, or simply argmax \cite{kim2020unsupervised} to transform network outputs into new pseudo-labels, GraPL uses these response vectors to define the unary energy of a patch-level MRF graph of the image.
    
    This approach for patch clustering introduces some advantages to our method. First, while the MRF modeling step is done primarily to impose a spatial coherence prior, due to the known shrinking bias of graph cuts \cite{kolmogorov2005metrics}, the resultant partition also smooths segment boundaries and reduces the number of distinct segments, leading to natural model selection. The spatial regularization introduced by the proposed graph can also be generalized to accommodate other classical graph formulations that consider segmentation seeds \cite{boykov2001fast}, appearance disparity \cite{tang2013grabcut}, curvature \cite{el2010fast}, or target distributions \cite{ayed2010graph}. Finally, in contrast to methods that discover objects by clustering patch embeddings arising from pretrained transformers and applying a segmentation head \cite{hamilton2022unsupervised} or CRF refinement \cite{wang2022tokencut, wang2023cut}, GraPL considers patch embeddings only as way to guide its training stage, yielding a final pixel-level segmentation map without postprocessing. We consider our graph cut-based approach to handle rich patch features beneficial, as we do not overly depend on their clustering power, and simply reference them as guidance when regularizing our training.

\section{Experiments}

    \subsection{Implementation  and Experimental Setup} \label{sec:exp_setup_implementation}
        \paragraph{Segmentation Task}
        To evaluate the performance and behaviors of GraPL, the algorithm was tasked with segmenting the 200 image test set of BSDS500 \cite{arbelaez2011bsds} using a variety of hyperparameters. Segmentation performance is measured in terms of mean intersection over union (mIoU) \cite{garciagarcia2017review}, with predicted segments matched one-to-one with target segments using a version of the Hungarian algorithm modified to accommodate $\hat{K} \neq  K^*$, where $\hat{K}$ is the number of distinct segments in the final segmentation, and $K^*$ is the number of segments in the ground truth. Results are averaged across 10 different random seeds for initialization.


        \paragraph{Hyperparameters}
        Unless otherwise specified, the following configuration was used during testing. Pseudo-labels were initialized according to the SLIC \cite{achanta2012slic} based algorithm described in Section \ref{para:initialization_experiment}. GraPL was run for four training iterations, and the number of gradient steps in the loss minimization at each iteration was 40, 32, 22, and 12 respectively. $K_0$, the number of initial segments was set at 14, and $d$ was set to 32. Graph cuts were implemented using pyGCO \cite{pygco}, and the pairwise energy coefficient, $\lambda$, was set to 64. The continuity loss was assigned a weight of $\mu=3$. The $L2$ norm between DINOv2 \cite{oquab2023dinov2} (ViT-L/14 distilled) patch embeddings was used as an affinity metric to determine pairwise weights. 

        \paragraph{Network Architecture}
        An intentionally minimal CNN architecture was used, consisting of 2 $3\times 3$ unpadded convolutional layers with 32 and 8 filters, respectively. In $F_\theta'$, this is followed by a dense classification head with $K_0$ units, and in $F_\theta$ it is followed by a $(\frac{h}{d} - 4) \times (\frac{w}{d} - 4)$ convolutional segmentation head with $K_0$ filters. The network layers are each separated by batch normalization, $\tanh$ activations, and dropout with rate 0.2. Without padding, our network is subject to certain regularization implications. In CNNs, zero padding an image has the effect of dropping out some of the weights of the subsequent convolutional layer. As our method requires the training phase to be executed on unpadded images, it is effectively deprived of this regularization feature. We found that applying dropout before the first convolutional layer all but resolved issues arising from the network's lack of padding. Despite its simplicity, this network is complex enough to achieve reliable segmentations, and more complex networks did not lead to better performance. In our work, we also abstain from using padding on our inference phase, which results in $\hat{S} = F_\theta(I)$ being of a size smaller than $\Omega$, due to the convolution operations in $F_\theta$. GraPL handles this discrepancy by interpolating $\hat{S}$ to the original dimensions via nearest neighbor interpolation.  Networks were implemented using PyTorch 2.0.1 \cite{pytorch}\footnote{A demo implementation of GraPL is available at \url{https://github.com/isaacwasserman/GraPL}}.

        \paragraph{Early Stopping}
        If a cross entropy loss of less than 1.0 was reached during the first iteration, it was stopped early, and new pseudo-labels were assigned. During the first iteration, we are fitting the initial pseudo-labels, which are either arbitrary or assigned by SLIC. By imposing this early stopping condition, we are avoiding the local minima where GraPL may be overfitting to a less performant (or worse, arbitrary) segmentation.

    \subsection{Ablation Studies}
    
        \paragraph{Initialization} \label{para:initialization_experiment}
            As an iterative algorithm, proper initialization is an important factor in training GraPL. Although similar deep clustering algorithms have used randomly initialized pseudo-labels \cite{caron2018deep, kim2020unsupervised}, we were unsure whether ignoring more principled approaches was leaving performance on the table. To answer this question, we compared four initialization strategies: ``patchwise random,'' ``seedwise random,'' ``spatial clustering,'' and an approach based on SLIC \cite{achanta2012slic}. The ``patchwise random'' approach individually assigns each patch $p$ in $\mathcal{P}$ a random label. In the ``seedwise random'' strategy, we select $K_0$ random patches and assign them each one of the $K_0$ labels; the remaining patches are assigned the label of the patch closest to them. For ``spatial clustering,'' patches are clustered using $k$-means according to their $(x,y)$ spatial coordinates to form $K_0$ clusters of roughly equal size. In the SLIC-based approach, we unfold a $K_0$ cluster SLIC segmentation with low compactness into the same patches as the input image. The onehot labels of these patches are averaged and normalized to produce soft labels. These soft initializations are an attempt to regularize and retain all salient features of the patch during training.

            

            \begin{figure}[b]
                \centering
                \input{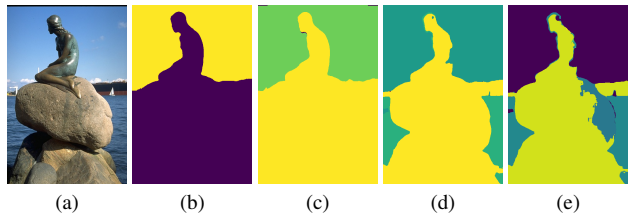}
                \caption{Example of undersegmentation from non-SLIC initialization. (a) Input image. (b) Patchwise Random ($\hat{K}=2$). (c) Seedwise Random ($\hat{K}=4$). (d) Spatial Clustering ($\hat{K}=4$). (e) SLIC ($\hat{K}=6$).}
                \label{fig:initialization_khat}
            \end{figure}
            
            Tests demonstrated that patchwise random initialization is not an ideal choice for GraPL (Table \ref{tab:initialization}). This is likely because it encourages a disregard for spatial coherence during the first and most important iteration. While SLIC was shown to be the best choice out of the methods tested, seedwise random and spatial clustering initialization performed only 1.0\% and 0.6\% worse, respectively, and the algorithm could likely be tuned such that they meet or exceed the performance of SLIC. However, in the current configuration, we notice a tendency for both of these methods to result in undersegmentation, in which $\Delta K = K_0 - \hat{K}$ is considerably higher than the SLIC version (Figure \ref{fig:initialization_khat}).

        \paragraph{Pairwise Edge Weights}
            The pairwise energy function (Eq. \ref{equation:pairwise_energy}) used by GraPL includes an affinity function $\text{aff}(p,q)$. Designed with vision transformers in mind, this function is defined by the Euclidean distance between some patch metric or embedding $m(p)$ for $p \in \mathcal{P}$.
            
            Though DINOv2 \cite{oquab2023dinov2} has been shown to produce excellent, fully unsupervised features on the patch level, requiring minimal supervised fine-tuning to produce an effective segmentation model \cite{oquab2023dinov2}. However, it's unclear whether the features are easily separable using unsupervised methods. 
            
            We examined the applicability of three definitions for $m(p)$: DINOv2 embedding, mean RGB color, and patch position. To produce the final DINO embeddings, images were resized to $14d \times 14d$, such that each GraPL patch corresponds to a $14 \times 14$ DINO patch. These embeddings were reduced to $K_0$ dimensions using 2nd degree polynomial PCA. As a baseline, we also tested a version where the fully connected graph was replaced with a 4-neighborhood lattice of uniformly weighted edges.
    
            In our tests, DINOv2 embeddings were a significantly better metric than distance alone (Table \ref{tab:graph_weights}). However, they were outperformed by simple RGB color vectors. Acknowledging DINOv2's ability to act as a feature extractor for supervised segmentation, further research is needed to determine what types of transformations are necessary for converting the embeddings into a better affinity metric.

            \noindent
            \begin{minipage}[t]{\linewidth}
                \noindent
                \begin{minipage}[b]{0.5\linewidth}
                \begin{table}[H]
                  \centering
                  \begin{tabular}{lc}
                    \toprule
                    Initializer       & mIoU  \\
                    \midrule
                    Patchwise Rand.   & 0.496 \\
                    Seedwise Rand.    & 0.507 \\
                    Spatial           & 0.509 \\
                    SLIC              & \textbf{0.512} \\
                    \bottomrule
                  \end{tabular}
                  \caption{Comparison of pseudo-label initialization methods}
                  \label{tab:initialization}
                \end{table}
                \end{minipage}
                \hfill
                \noindent
                \begin{minipage}[b]{0.46\linewidth}
                \begin{table}[H]
                  \centering
                  \begin{tabular}{lc}
                    \toprule
                    Metric     & mIoU  \\
                    \midrule
                    Uniform             & 0.459 \\
                    Position            & 0.476 \\
                    Color               & \textbf{0.527} \\
                    DINOv2              & 0.512 \\
                    \bottomrule
                  \end{tabular}
                  \caption{Comparison of pairwise weighting metrics}
                  \label{tab:graph_weights}
                \end{table}
                \end{minipage}
            \end{minipage}

\begin{figure}
                \centering
                \begin{tikzpicture}
  \begin{axis}[
      tick align=outside,
      tick pos=left,
      xlabel={Gradient Step},
      xmin=0, xmax=115,
      ymin=0, ymax=3.4,
      ylabel={Cross Entopy},
      ylabel near ticks,
      xtick={0, 40, 72, 94},
      xlabel near ticks,
      ticklabel style = {font=\small},
      minor x tick num=4,
      ytick style={color=black},
      width=\linewidth,
      height=0.55\linewidth,
      legend style={at={(0.27,0.35)}, font=\small},
    ]
    \addplot [semithick]
    table {%
        0 2.9239849984645843
        1 3.002001852989197
        2 3.390752147436142
        3 3.200280270576477
        4 2.7946790713071823
        5 2.6275452011823655
        6 2.511121305823326
        7 2.3340234225988388
        8 2.158817562460899
        9 2.0834860122203827
        10 1.9576450657844544
        11 1.8632868993282319
        12 1.787639085650444
        13 1.7101116240024568
        14 1.6556679838895798
        15 1.6047415298223495
        16 1.579364545047283
        17 1.52685662150383
        18 1.5147354629635812
        19 1.473216785490513
        20 1.4484445771574974
        21 1.4359987324476242
        22 1.4102113509178162
        23 1.4027864649891852
        24 1.3837179002165794
        25 1.3674263980984689
        26 1.3591498869657517
        27 1.3602891391515732
        28 1.350519226193428
        29 1.3347212034463882
        30 1.332302292883396
        31 1.3123891299962998
        32 1.3107414489984512
        33 1.303725138604641
        34 1.305188876390457
        35 1.2901571932435036
        36 1.2900044757127762
        37 1.275952881872654
        38 1.2719578778743743
        39 1.2672333917021752
        40 1.3374231693148613
        41 0.8777504079043865
        42 0.7498509711772203
        43 0.6893068016320467
        44 0.6305874977260828
        45 0.5908575589954853
        46 0.5607393290102481
        47 0.5330341252684593
        48 0.5141757355630397
        49 0.4894430509209633
        50 0.48163396153599025
        51 0.47142834089696406
        52 0.4549091985076666
        53 0.4517375028505921
        54 0.4451264300942421
        55 0.437356695458293
        56 0.4291552801430225
        57 0.4210680088028312
        58 0.4290970441699028
        59 0.42086738187819717
        60 0.41727157577872276
        61 0.4051391928270459
        62 0.40641504634171727
        63 0.40652246758341787
        64 0.4054489202424884
        65 0.3978022265434265
        66 0.3924556789547205
        67 0.396591619476676
        68 0.38862720236182213
        69 0.3912436306104064
        70 0.3853556392341852
        71 0.3846201119571924
        72 0.32177659804932773
        73 0.2570112993195653
        74 0.24002149438485504
        75 0.22757017896510662
        76 0.22476911346428097
        77 0.2242235780134797
        78 0.21240446873009206
        79 0.21301912387833
        80 0.2131352021358907
        81 0.2091529713757336
        82 0.2085779482871294
        83 0.20103655504062773
        84 0.19688289252109825
        85 0.20493048403412104
        86 0.19912919167429208
        87 0.19440280443057417
        88 0.19598746688105165
        89 0.19533231060951947
        90 0.19044957228004933
        91 0.19161531994119285
        92 0.18893657154403626
        93 0.19028177415020764
        94 0.24975671273365152
        95 0.17619854608085006
        96 0.16065330002456904
        97 0.15838140335399659
        98 0.16011760702123864
        99 0.15902104049688204
        100 0.15422544127563015
        101 0.15352566142042634
        102 0.15428760170878378
        103 0.1559708119323477
        104 0.16012424374232068
        105 0.14880834370735102
        106 0.1497641408792697
        107 0.14542144284409006
        108 0.15006059573352104
        109 0.15002961178703117
        110 0.14729703046905343
        111 0.1435992850526236
        112 0.14137395695986923
      };
      \addplot [semithick, dashed]
        table {%
            0 2.9239849984645843
            1 3.002001852989197
            2 3.390752147436142
            3 3.200280270576477
            4 2.7946790713071823
            5 2.6275452011823655
            6 2.511121305823326
            7 2.3340234225988388
            8 2.158817562460899
            9 2.0834860122203827
            10 1.9576450657844544
            11 1.8632868993282319
            12 1.787639085650444
            13 1.7101116240024568
            14 1.6556679838895798
            15 1.6047415298223495
            16 1.579364545047283
            17 1.52685662150383
            18 1.5147354629635812
            19 1.473216785490513
            20 1.4484445771574974
            21 1.4359987324476242
            22 1.4102113509178162
            23 1.4027864649891852
            24 1.3837179002165794
            25 1.3674263980984689
            26 1.3591498869657517
            27 1.3602891391515732
            28 1.350519226193428
            29 1.3347212034463882
            30 1.332302292883396
            31 1.3123891299962998
            32 1.3107414489984512
            33 1.303725138604641
            34 1.305188876390457
            35 1.2901571932435036
            36 1.2900044757127762
            37 1.275952881872654
            38 1.2719578778743743
            39 1.2672333917021752
            40 2.930468307733536
            41 2.984072577953339
            42 1.8696138766407966
            43 1.6609648269414903
            44 1.505679472386837
            45 1.3662208250164987
            46 1.2456979498267173
            47 1.1543793085962535
            48 1.1130178099870682
            49 1.0296160317957401
            50 0.9620708007365465
            51 0.9079866857826709
            52 0.8682579904794693
            53 0.8262513865530491
            54 0.7650364865362644
            55 0.7080706070363522
            56 0.6870394607633352
            57 0.6438274524360895
            58 0.6303686109557748
            59 0.6057106920331716
            60 0.5880108394473791
            61 0.5790491247177124
            62 0.5511183752119542
            63 0.547877748273313
            64 0.5310503500699997
            65 0.5147321688383818
            66 0.516669618524611
            67 0.4985682263970375
            68 0.4865213795751333
            69 0.47356356747448447
            70 0.47409215502440927
            71 0.4677237793058157
            72 2.9301608347892762
            73 2.8375734603405
            74 1.570602160990238
            75 1.425111635476351
            76 1.2432054777443409
            77 1.0877655009925364
            78 0.9901908230036497
            79 0.8979609034955501
            80 0.8103548483550549
            81 0.7378229263424874
            82 0.6661148311197758
            83 0.5963551821187139
            84 0.546315360236913
            85 0.5048200654610991
            86 0.46516365229152146
            87 0.4273251502774656
            88 0.3908598465670366
            89 0.37269552418234525
            90 0.3530880012916168
            91 0.3463592603200232
            92 0.32188455085270107
            93 0.3151652771602676
            94 2.929662219285965
            95 2.9206397971510887
            96 1.7065516763925552
            97 1.4559365421533585
            98 1.2907290095835924
            99 1.2282614988833667
            100 1.0411707570403814
            101 0.9385415641963482
            102 0.8266394575685263
            103 0.758269913867116
            104 0.6749462686106562
            105 0.609265960752964
            106 0.5401237007975578
            107 0.47884578075259926
            108 0.4333362998696975
            109 0.39357720929430795
            110 0.3436130968562793
            111 0.31027726398402594
            112 0.302083367932064
        };
        \legend{Warm, Cold}
  \end{axis}
\end{tikzpicture}
                \caption{Comparison of loss curves using warm and cold starting methods, averaged over all test images in BSDS500 \cite{arbelaez2011bsds}. Here we consider the loss value at the end of each gradient step. On the $x$-axis, we depict the instants where a new training iteration starts.}
                \label{fig:warm_cold_start}
            \end{figure}
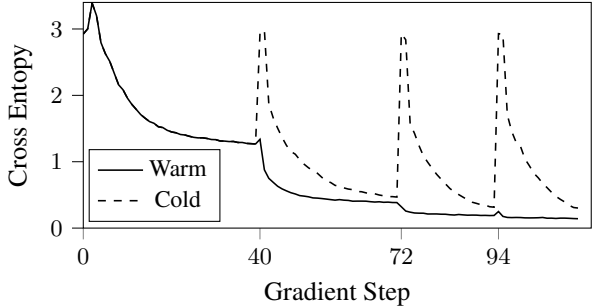

        \paragraph{Warm Start}
            GraPL is designed to train the same network continuously throughout all iterations. This is in contrast to similar iterative methods which prefer a ``cold start,'' re-initializing the parameters of the surrogate function prior to subsequent iterations. Preliminary tests showed that in our case, a ``warm start'' approach is preferred to re-initializing the network each time. These two approaches in fact produce very different loss curves (Figure \ref{fig:warm_cold_start}). Cold starts produce large spikes in loss at the beginning of each training iteration, whereas warm starts require only minor adjustments at these points. We expect that the first iterations of training provide important feature learning to the first layers of the network. By starting cold at each iteration, subsequent iterations are unable to benefit from the learned low-level feature detectors and therefore present a more unstable training phase.
            
        \paragraph{Pairwise Energy Coefficient ($\lambda$)}
            GraPL uses graph cuts to generate each new set of pseudo-labels, working on the theory that this graphical representation of the image provides an important spatial coherence prior, which is perhaps missing from similar unsupervised methods, and accounts for its success. Furthermore, GraPL relies on pairwise costs as well as the continuity loss to gradually decrease $\hat{K}$. To test these ideas, we evaluated the segmentation performance of the algorithm as well as $\Delta K$ across different values of $\lambda$, the hyperparameter which defines the scale of the pairwise energy as defined in Eq. \ref{equation:pairwise_energy}.
            
            When $\lambda=0$, cutting any non-terminal edge incurs no cost. In this case, the function of the cut is effectively the same as the argmax clustering step found in \cite{kim2020unsupervised}, as pseudo-labels are entirely dependent on the current network response vectors. As $\lambda$ increases, network response vectors are made less influential in the pseudo-label assignment process, as expected.
            
            \begin{figure}
              \centering
              \begin{subfigure}{.5\linewidth}
                \centering
                \begin{tikzpicture}
                \node at (0.25,1.2) {\textbf{(a)}};
                  \begin{axis}[
                      log basis x={2},
                      tick align=outside,
                      tick pos=left,
                      xlabel={$\lambda$},
                      xmin=0.5, xmax=256,
                      xmode=log,
                      xtick={0.5, 1, 2, 4, 8, 16, 32, 64, 128, 256},
                      xticklabels={0, , 2, , 8, , 32, , 128, },
                      ylabel={mIoU},
                      ymin=0.46, ymax=0.52,
                      ytick style={color=black},
                      width=4.2cm,
                      height=3cm,
                    ]

                    \addplot [semithick]
                    table {%
                        0.5 0.475594526580893
                        1 0.48541679090677
                        2 0.489487170399904
                        4 0.494947384173972
                        8 0.501129224425331
                        16 0.504353126090412
                        32 0.509232860180572
                        64 0.513311869927415
                        128 0.512401978848862
                        256 0.510680382646683
                      };
                  \end{axis}
                \end{tikzpicture}
                \label{fig:lambda_miou}
              \end{subfigure}%
              \begin{subfigure}{.5\linewidth}
                \centering
                \begin{tikzpicture}
                \node at (0.25,1.2) {\textbf{(b)}};
                  \begin{axis}[
                      log basis x={2},
                      tick align=outside,
                      tick pos=left,
                      xlabel={$\lambda$},
                      xmin=0.5, xmax=256,
                      xmode=log,
                      xtick={0.5, 1, 2, 4, 8, 16, 32, 64, 128, 256},
                      xticklabels={0, , 2, , 8, , 32, , 128, },
                      ylabel={$\Delta K$},
                      ymin=11.4, ymax=12.2,
                      width=4.2cm,
                      height=3cm,
                    ]
                    \addplot [semithick]
                    table {%
                        0.5 11.429
                        1 11.5965
                        2 11.6765
                        4 11.77
                        8 11.8485
                        16 11.9885
                        32 12.019
                        64 12.15
                        128 12.176
                        256 12.0525
                      };
                  \end{axis}
                \end{tikzpicture}
                \label{fig:lambda_deltak}
              \end{subfigure}
              \vspace{-30pt}
              \caption{Effects of pairwise energy coefficient. (a) Effect of $\lambda$ on mIoU. (b) Effect of $\lambda$ on $\Delta K$.}
              \label{fig:lambda_tests}
            \end{figure}
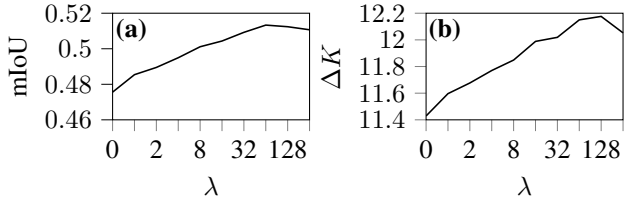
            The results in Figure \ref{fig:lambda_tests} demonstrate a logarithmic increase in segmentation performance as $\lambda$ is raised from $0$ through $64$. However, increasing $\lambda$ to values higher than 64 tends to result in comparatively poor performance. Because increasing $\lambda$ strengthens pairwise connections, we would expect it to be closely correlated with $\Delta K$. When $\lambda \leq 64$, we observe this behavior; however, higher values actually result in a plateau or slight decrease in $\Delta K$.
            
            In a configuration where the pairwise edges were uniformly weighted (or weighted according to spatial distance), we would expect higher than optimal values of $\lambda$ to push $\Delta K$ too high and produce oversimplified segmentations, where multiple target segments are collapsed into a single predicted segment. However, when pairwise edges are weighted by patch encoder embedding affinity, pushing $\lambda$ too high can instead result in an overly detailed segmentation, in which the graph cut considers the pairwise energy (dictated by the patch embeddings) more than the unary weights learned by GraPL.

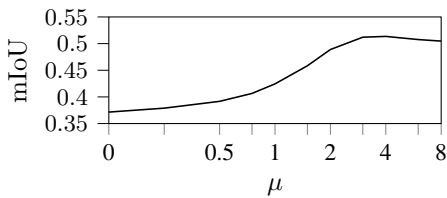
\begin{figure}[b]
              \centering
                \centering
                \hspace*{-1cm}
                \begin{tikzpicture}
                  \begin{axis}[
                      log basis x={2},
                      tick align=outside,
                      tick pos=left,
                      xlabel={$\mu$},
                      xmin=0.125, xmax=8,
                      xmode=log,
                      ticklabel style = {font=\small},
                      xtick={0.125, 0.25, 0.5, 0.75, 1, 1.5, 2, 3, 4, 6, 8},
                      xticklabels={0, , 0.5, , 1, , 2, , 4, , 8},
                      ylabel={$\operatorname{mIoU}$},
                      ymin=0.35, ymax=0.55,
                      ytick style={color=black},
                      width=6cm,
                      height=3cm,
                    ]
                    \addplot [semithick]
                    table {%
                        0.125 0.37154687607722253
                        0.25 0.37889801417901375
                        0.5 0.3916806735791856
                        0.75 0.40651565094956404
                        1 0.4244394644625914
                        1.5 0.4584265297494407
                        2 0.4888776007687718
                        3 0.5120218815581796
                        4 0.5134930146702269
                        6 0.5075909862585035
                        8 0.5046944848315008
                      };
                  \end{axis}
                \end{tikzpicture}
              \vspace{-10pt}
              \caption{Effect of spatial continuity loss weight on mIoU}
              \label{fig:continuity_miou}
            \end{figure}

        \paragraph{Continuity Loss}
            Spatial continuity loss, first introduced in \cite{kim2020unsupervised}, provides GraPL a spatial coherence prior which penalizes the network directly at each gradient step, rather than through the graph cut produced pseudo-labels at the end of each iteration. Though shown effective in \cite{kim2020unsupervised}, we
            instinctively believed that it would be redundant in a graphically guided pipeline like GraPL. However, we observed that the combination of the two different spatial coherence priors produced more accurate segmentations than either one alone (Figure \ref{fig:continuity_miou}).

            In practice, we noted that this loss has a different mechanism of action than the graphical coherence prior. In the absence of this spatial loss, GraPL employs a level of trust in the patch encoder that may be unfounded, as the pairwise energy only penalizes the separation of patches with a great affinity; but when using a patch encoder like DINO, which is defined by a large neural network, the edge weights may be high variance. In this case, increasing $\lambda$ only serves to emphasize the patch encoder's bias for certain edges. However, increasing the weight of the spatial continuity loss applies a higher penalty for \emph{all} edges. In effect, it could be compared to an additive bias term in the pairwise energy function that raises the cost, no matter the patch affinity.
    
        \paragraph{Patch Size}
            Patch-based approaches are faced with a choice between granularity (with smaller patches) and the information richness of input (with larger patches). In GraPL's case, smaller patches also entail more complex graphs that take longer to solve, and larger patches entail higher memory usage. We found that setting $d$ equal to 32 offered both optimal performance and near optimal efficiency (Table \ref{tab:d}).
            
            \begin{table}[ht]
              \centering
              \begin{tabular}{lcc}
                \toprule
                $d$    & mIoU  & Seconds per Image\\
                \midrule
                8  & 0.248 & 3.49 \\
                16 & 0.372 & \textbf{1.72} \\
                32 & \textbf{0.512} & 1.75 \\
                64 & 0.496 & 6.98 \\
                \bottomrule
              \end{tabular}
              \caption{mIoU and segmentation time as a function of patch size. Time measurements are based on segmentation of BSDS500 test set on an Nvidia A100 GPU and AMD Epyc 7343 CPU.}
              \label{tab:d}
            \end{table}

        \begin{figure*}
            \centering
            \input{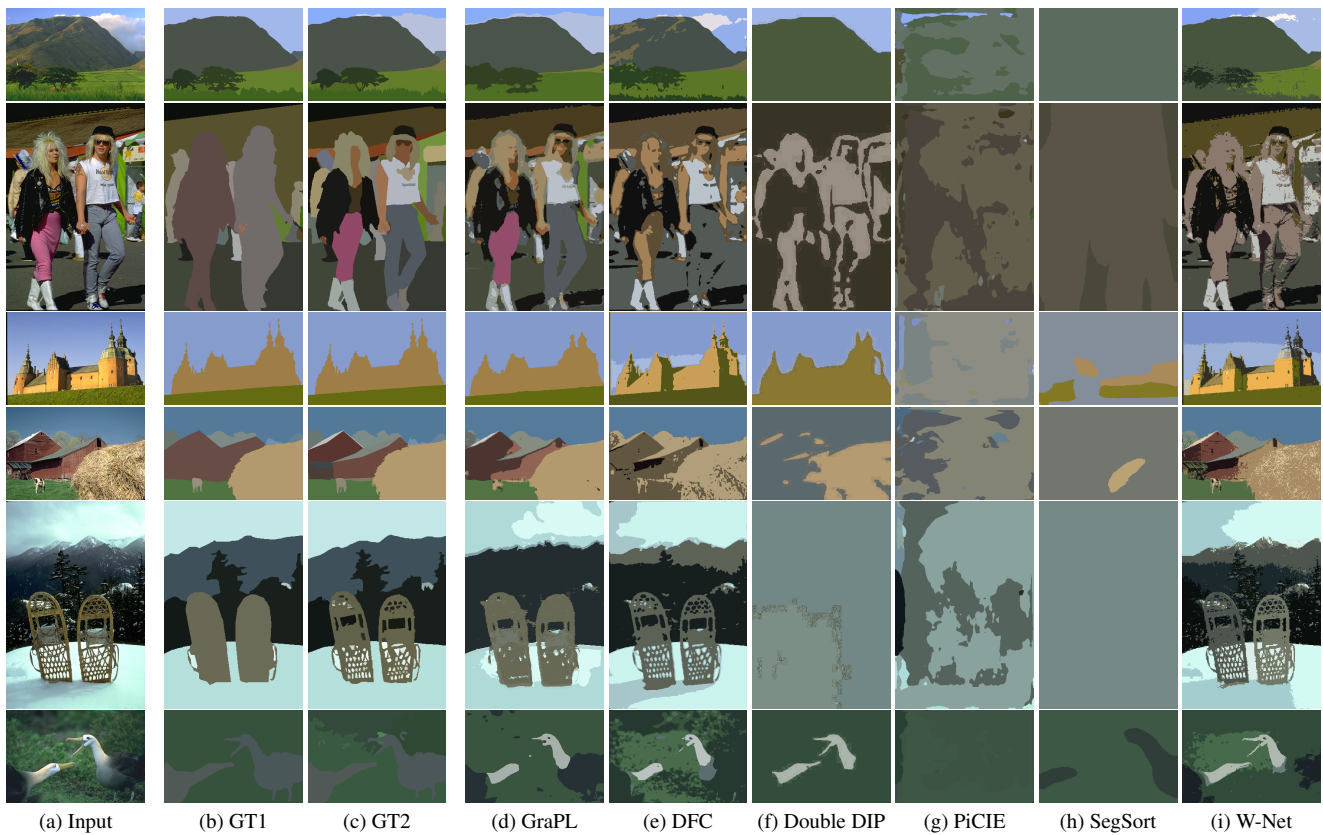}
            \caption{Qualitative comparison of GraPL to other deep learning-based  unsupervised segmentation methods. We selected two of the available ground truth segmentations (GT1 and GT2) for comparison, one more detailed and one less detailed.}
            \label{fig:qualitative_comparison}
        \end{figure*}
        
    \subsection{Comparison to Other Methods}
        We compared the segmentation performance of GraPL to six other unsupervised deep-learning methods: Differentiable Feature Clustering (DFC) \cite{kim2020unsupervised}, Invariant Information Clustering (IIC) \cite{ji2019invariant}, Pixel-level feature Clustering
        using Invariance and Equivariance (PiCIE) \cite{cho2021picie}, Segment Sorting (SegSort) \cite{hwang2019segsort} and W-Net \cite{xia2017w}. We also tested two baselines which use SLIC to segment images based on RGB and DINOv2 \cite{oquab2023dinov2} patch embeddings (interpolated to the pixel level). These baselines were selected to demonstrate that the success of our method does not simply originate from its initialization or its pretrained guidance. PiCIE and SegSort  were trained on their preferred datasets, COCO-Stuff \cite{caesar2018coco} and PASCAL VOC 2012 \cite{everingham2012pascal}, as BSDS500 is too small, while the others used only BSDS500.
        
        Table \ref{tab:quantitative_comparison} summarizes the quantitative comparative results of the above methods, where segmentation performance was measured in terms of both mIoU and pixel accuracy \cite{garciagarcia2017review}. As with mIoU, pixel accuracy was computed using a one-to-one label matching strategy. Figure \ref{fig:qualitative_comparison} displays some segmentation results from the above methods for qualitative comparison.

        
        \begin{table}[ht]
            \centering
            \begin{tabular}{lcc}
                \toprule
                Method & mIoU & Accuracy  \\
                \midrule
                SLIC (RGB features)  & 0.137 & 0.416  \\
                SLIC (DINOv2 features) & 0.258 & 0.280  \\
                DFC \cite{kim2020unsupervised}        & 0.398          & 0.505                    \\
                DoubleDIP \cite{gandelsman2019double} & 0.356          & 0.423                    \\
                IIC \cite{ji2019invariant}\dag        & 0.172          &                               \\
                PiCIE \cite{cho2021picie}             & 0.325          & 0.405                    \\
                SegSort \cite{hwang2019segsort}       & 0.480          & 0.505                    \\
                W-Net \cite{xia2017w}                 & 0.428          & 0.531           \\
                GraPL (proposed)                          & \textbf{0.527} & \textbf{0.569}           \\
                \bottomrule
            \end{tabular}
            \caption{BSDS500 performance comparison of GraPL with other unsupervised deep-learning methods and baselines. \\\dag The value listed for IIC is sourced from \cite{kim2020unsupervised}.}
            \label{tab:quantitative_comparison}
        \end{table}
        
        Compared to other unsupervised methods, GraPL is able to decompose complex foregrounds into detailed yet semantically salient components. Notice how GraPL is able to pick up on small details like sunglasses in the distance while ignoring less relevant features of the image, such as creases in clothing. In many cases, it is able to handle color gradient variation, usually present in sky backgrounds or shadow regions. On occasion, GraPL detects segments that are not present in the ground-truth, such as the bird heads on the last qualitative example, which, despite being reasonable, decreases its quantitative performance. Finally, it also struggles to detect fine structures, such as castle tops, small holes and bird beaks. Despite that, our proposed method is able to outperform all of the compared methods by a margin of at least 6.9\% in accuracy and 9.3\% mIoU. It is also worth noting the low performance of our baselines, when compared to GraPL. This demonstrates that our method does not merely rest on the success of our our initializer, SLIC. Instead, GraPL's success is a product of its novel training and inference methodology.
    
        \section{Conclusion}
        In this paper, we introduce GraPL, a deep learning-based unsupervised segmentation framework that approaches the problem by solving a patch clustering surrogate task to learn network parameters which are then used for pixel-level classification. Additionally, GraPL is the first deep learning method to employ a graph cut regularizer during training, which encourages spacial coherence and leverages the discriminative power of patch embeddings. Furthermore, it seamlessly translates patch-level learning to the pixel-level without the need for postprocessing. Our experiments demonstrate our algorithm's promising capacities, as it is able to outperform many state-of-the-art unsupervised segmentation methods. Our work can be seen as further evidence for the benefit of using graph cuts in deep learning, especially in the context of unsupervised segmentation. 

{\small
\bibliographystyle{ieee_fullname}
\bibliography{egbib}
}

\end{document}